\documentclass[journal]{IEEEtran}
\usepackage{amsmath, amsfonts, graphicx, amssymb}
\usepackage{algpseudocode, algorithm, booktabs, color}

\usepackage{multirow}

\def\h{\mathbf{h}}
\def\H{\mathbf{H}}

\def\bR{\mathbb{R}}

\def\w{\mathbf{w}}
\def\W{\mathbf{W}}
\def\x{\mathbf{x}}
\def\X{\mathbf{X}}
\def\y{\mathbf{y}}
\def\Y{\mathbf{Y}}
\def\z{\mathbf{z}}
\def\Z{\mathbf{Z}}

\newcommand{\one}{\boldsymbol{1}}
\newcommand{\calP}{\mathcal{P}}

\begin{document}

% \title{Non-negative Matrix Completion on Triple Simplex for Project Expense Forecasting}
\title{Triple Simplex Matrix Completion for Expense Forecasting}

\author{Cheng Qian,~\IEEEmembership{Member,~IEEE,}
        Lucas M. Glass,
        and Nicholas D. Sidiropoulos,~\IEEEmembership{Fellow,~IEEE}
        
\thanks{C. Qian and L. Glass are with the Analytics Center of Excellence, IQVIA, USA. E-mail: alextoqc@gmail.com (C. Qian), lmglass@us.imshealth.com (L. Glass)}
\thanks{N. Sidiropoulos is with the Department of Electrical and Computer Engineering, University of Virginia, Charlottesville, VA 22904. E-mail: nikos@virginia.edu}
}

\maketitle

\begin{abstract}

Forecasting project expenses is a crucial step for businesses to avoid budget overruns and project failures. Traditionally, this has been done by financial analysts or data science techniques such as time-series analysis. However, these approaches can be uncertain and produce results that differ from the planned budget, especially at the start of a project with limited data points. This paper proposes a constrained non-negative matrix completion model that predicts expenses by learning the likelihood of the project correlating with certain expense patterns in the latent space. The model is constrained on three probability simplexes, two of which are on the factor matrices and the third on the missing entries. Additionally, the predicted expense values are guaranteed to meet the budget constraint without the need of post-processing. An inexact alternating optimization algorithm is developed to solve the associated optimization problem and is proven to converge to a stationary point. Results from two real datasets demonstrate the effectiveness of the proposed method in comparison to state-of-the-art algorithms.
\end{abstract}

\begin{IEEEkeywords}
Matrix factorization, missing imputation, probability simplex, revenue prediction, expense forecast.
\end{IEEEkeywords}
\section{Introduction}

% Predicting labor and non-labor costs at the project level is an important task in project management. An accurate expense forecast model can help the product team forecast the resources needed in each stage of the project and make sure the project stays on track. In certain businesses, such as contract research companies, revenue is closely related to costs, since a large part of the contract awards come from the customer reimbursing expenses.

% In many organizations, financial analysts are typically responsible for forecasting project expenses by examining the costs of similar past projects. However, due to variations in factors such as budget and project type, these estimates may not always be reliable. To improve expense forecasting, data science techniques can be utilized. Since project expenses follow a time-series pattern, approaches like the auto-regressive integrated moving average (ARIMA) algorithm can be applied to calculate expenses. However, ARIMA may not be suitable for budget-related issues and may be particularly inaccurate in the early stages of a project. Recurrent neural networks, such as long short-term memory networks (LSTM), are typically not effective for project-level expense forecasting since there may not be sufficient data for training and budget constraints may not be adequately incorporated.

The accurate forecasting of labor and non-labor costs at the project level is a crucial aspect of effective project management. A reliable expense forecast model can aid project teams in anticipating the resources required at each stage of the project, ensuring timely completion and adherence to budget constraints. This is particularly important in businesses where revenues are closely linked to costs, such as contract research organizations, where a significant portion of contract awards is derived from customer reimbursements for expenses incurred.

Traditionally, financial analysts within organizations have relied on historical cost data from similar projects to predict future project expenses. However, these estimations can be prone to errors due to variations in factors such as budget and project type, leading to suboptimal decision-making and potential financial losses. To enhance the accuracy of expense forecasting, the application of advanced data science techniques has gained increasing prominence in recent years.

Project expenses typically exhibit time-series patterns, making them amenable to analysis using time-series forecasting techniques. One such approach is the auto-regressive integrated moving average (ARIMA) model, which has shown promise in predicting project expenses. However, ARIMA is not the most suitable method for addressing budget-related concerns and may exhibit reduced accuracy in the early stages of a project. An alternative approach involves the use of recurrent neural networks (RNNs), such as long short-term memory networks (LSTMs) \cite{hochreiter1997long}, which have demonstrated success in various time-series prediction tasks. However, the application of LSTMs to project-level expense forecasting is not without challenges. The scarcity of training data and lack of disciplined ways of capturing budget constraints limit the effectiveness of LSTM-based models in this context.

In light of these limitations, there is a pressing need for the development of novel expense forecasting models that can effectively leverage available data, account for project-specific constraints, and ultimately improve the accuracy of cost predictions. The integration of cutting-edge techniques from the fields of signal processing, machine learning, and optimization offers promising avenues for advancing the state-of-the-art in expense forecasting, paving the way for more efficient and cost-effective project management across a wide range of industries. 

This study aims to contribute to this ongoing research by proposing and investigating a novel approach to expense forecasting, with a particular focus on addressing the aforementioned challenges and enhancing the practical applicability of these techniques in real-world settings. In contrast to conventional time-series based methods, our proposed technique transforms the forecasting problem into a matrix completion problem with constraints. We introduce the Triple Simplex Matrix Completion (TSMC) algorithm, which utilizes three simplex projections to limit the factor matrices and missing values, ensuring that forecasts do not exceed budget constraints. Our method employs an inexact projected gradient descent algorithm to solve the non-negative matrix completion problem with three simplex constraints. Experimental results using two real datasets demonstrate that our proposed approach outperforms existing algorithms in expense prediction.

\section{Problem Statement}
Assume that there are $N$ projects, out of which $K$ (where $0 < K < N$) are completed and the rest are ongoing. The objective is to estimate the future expenses for the ongoing projects over time, while adhering to budget constraints. However, there are several difficulties that make this problem challenging. First, the length of each project varies, ranging from a few months to several years. Second, since the budgets for different projects may differ significantly, appropriately scaling the cost values across diverse projects presents a significant challenge. Third, some projects have specific end dates, while others do not. Determining how to incorporate the budget and targeted end dates (TEDs) optimally poses a challenge in modeling the expense forecasting problem.
\section{Proposed Method}

% \subsection{Motivation}\label{section:motivation}
% In practice, depending on the types of projects, some of them progress quickly, while others may progress slowly until their budgets are burned out. Assume there are a fixed number of expense patterns, each of which can be fast, moderate, or slow and describes the progression of a project within a specific planned budget. Then the progression of a project can be approximated using a combination of the expense patterns.

%As we will see later, such an assumption makes a lot of sense in practice and will not affect the expense prediction of projects that are progressing fast, especially for those with a specific TED. For example, giving $M=12$ months, when the TED of a new project is six months, its budget will be distributed over the first six months, and the expenses from the seventh to the twelfth are all zero.
We assume that the maximum duration of any project is $M$ time units (e.g., months). Parameter $M$ depends on the type of projects considered, and can be determined from historical project data. Let $\y_n=[y_{1n}, \cdots, y_{Mn}]^T\in\bR^M$ be the expense vector of the $n$-th project, where $y_{mn}\geq 0$ denotes the $m$-th non-negative expense value. We consider the unobserved expenses as missing data in $\y_n$. We add trailing zeros to bring the duration of each project to $M$. For ongoing projects, we only add zeros after their respective TEDs if these TEDs are known beforehand, e.g., when there is a contractual end date. 

By stacking $N$ projects into a matrix, we have
\begin{align}\label{eq:Y}
    \Y = \begin{bmatrix}
        \y_1 & \y_2 & \cdots & \y_N
    \end{bmatrix} \in \bR^{M\times N}.
\end{align}
%Define $\Omega$ as the index set that contains all missing values in $\Y$. 
Now the task of estimating expenses can be thought of as filling in the missing values in $\Y$. Experience with real project expense data suggests that projects with similar characteristics tend to have similar expense patterns. For instance, smaller projects typically have faster expense patterns, while larger projects tend to have slower expense patterns that take years to complete. Thus, it is reasonable to assume that the expense pattern of any project can be approximated using a linear combination of a limited number of expense patterns, which implies that $\Y$ is low rank. We therefore propose to formulate the expense forecasting problem as a matrix completion problem subject to appropriate constraints. The constraints here are key to the success of our endeavor, as we will see.  

% With the above observations, we propose to model the expense forecast problem as a constrained matrix completion problem, where the goal is to impute the missing expense values in $\Y$ under certain constraints. One naive formulation can be
% \begin{equation}\label{eq:mc0}
% \begin{aligned}
%     \min_{\tilde{\Y}} ~&\mathrm{rank}(\tilde{\Y}) \\
%     \mathrm{s.~t.}~&\calP_\Omega(\Y) = \calP_\Omega(\tilde{\Y}) % \|\M \circledast (\Y - \W\H^T)\|_F^2
% \end{aligned}
% \end{equation}
% where $\mathrm{rank}(\cdot)$ is the matrix rank, $\calP_{\Omega}$ is the operation that only keeps the entries from index set $\Omega$ and zero-out the remaining entries, and $\Omega$ contains the index of all known entries in $\tilde\Y$.
%where $\W\in\bR^{M\times F}$ denotes the expense embedding matrix with its $m$-th row being the expense embedding at time-$m$ and its $n$-th column being the $n$-th expense pattern, $\H\in\bR^{N\times F}$ denotes the project embedding of , $\circledast$ is element-wise product and $\|\cdot\|_F$ is the Frobenius norm. Here, $F$ is the rank of $\Y$, i.e., the number of expense patterns in the dataset. However, \eqref{eq:mc0} is not the optimal way for expense prediction. The main reason is that it does not constrain the TED and budget information, which will ultimately produce expense estimates inconsistent with the budget constraint.

\subsection{Modeling}
There are three overarching constraints that must be considered in project expense forecasting practice: the known TED of the project, the allocated project budget, and the non-negativity of expense values. To ensure that expenses for each project are in agreement within the allocated budget we must enforce that $\sum_{m=1}^M y_{mn} = B_n$. However, order-of-magnitude differences in project budgets can cause problems in forecasting expenses for small and medium-size projects. A good way to address the scaling issue is to transform the absolute expense values to relative ones, normalized by the respective project's budget. Specifically, we can define $\x_n = \y_n / B_n$, where $\x_n$ is the expense vector expressed in terms of  fractions of the budget allocated to the $n$-th project.

Next, we enforce two constraints on $\x_n$. First, we require that the sum of its elements equals 1, which implies that the budget is fully utilized, as expected. Second, since expenses are non-negative, we need to ensure that each element of $\x_n$ is non-negative as well. Formally, we have
\begin{align}\label{eq:1xn}
    \one^T\x_n = 1,~ \x_n\geq\boldsymbol{0}
    %\sum_{m=1}^M x_{mn} = 1, ~x_{mn} \geq 0
\end{align}
where $\one=[1,\cdots,1]^T\in\bR^M$. 
%\reminder{From Nikos: I think that $\geq$ is normally used to denote positive semidefinite, whereas $\geq$ is used for element-wise non-negative. I leave it up to you if you want to change this.} -> Agreed!

With the assumption that $\Y$ has a low rank, we propose to model $\x_n$ as 
\begin{align}\label{eq:xn}
    \x_n = \W\h_n
    % x_{mn} = \w_m^T \h_n
\end{align}
where $\W=[\w_1,\cdots,\w_F]\in\bR^{M\times F}$ is a matrix containing a learned dictionary of expense patterns, with $\w_f$ being the $f$-th pattern, and $\h_n$ is the project embedding. Here, $F < \min(M, N)$ represents the (non-negative) matrix rank and it corresponds to the minimum number of components (expense pattern profiles) needed to model the data.

Since expenses are non-negative, we impose non-negativity constraints on $\W$ and $\h_n$. This means that the matrix $\W$ containing the prototype expense patterns and the project embedding $\h_n$ are both non-negative, denoted as $\W\geq\boldsymbol{0}$ and $\h_n\geq\boldsymbol{0}$.
By plugging in \eqref{eq:xn} into \eqref{eq:1xn}, we obtain the constraint that the sum of the estimated expenses should equal 1, which can be written as
\begin{align}\label{eq:wh}
    \one^T\W\h_n = 1, \W\geq \boldsymbol{0}, \h_n\geq \boldsymbol{0}.
\end{align}
Without loss of generality, we may assume that the sum of each column of $\W$ is 1, i.e., 
\begin{align}\label{eq:constW}
    \one^T\W = \one^T,~\W\geq\boldsymbol{0}
\end{align} 
%with the interpretation that the total expense in the latent space equals the budget constraint.
We can obtain the second constraint for our model by substituting \eqref{eq:constW} into \eqref{eq:wh}, i.e., $\one^T\h_n = 1, \h_n\geq 0$. %This constraint states that for each project, the sum of the elements in its corresponding embedding vector $\h_n$ should be equal to 1, and all elements in $\h_n$ should be non-negative. 
Interestingly,  the constraint of $\h_n$ takes the form of a probability vector, where each element represents the likelihood that the project belongs to a particular expense pattern. In matrix form, this constraint can be written as 
\begin{align}\label{eq:consth}
    \H\one = \one,~\H\geq\boldsymbol{0}
\end{align}
where $\H = [\h_1,\cdots,\h_N]^T \in\bR^{N\times F}$ and all elements in $\H$ are non-negative.

%\reminder{Notice I fixed the transposes inside [the brackets], which were wrong I think.} -> Agreed!

It is important to note that  the constraints in \eqref{eq:constW} and \eqref{eq:consth} only ensure that the low-rank approximation $\W\H^T$ satisfies the budget constraints. They do not guarantee that the sum of predicted and observed expenses for each project will also adhere to  the budget constraint. To address this issue, an additional constraint is required for the missing entries.
Let $\z_n$ represent the estimated expense vector for project-$n$. For each observed expense, we have $\calP_{\Omega}(\z_n) = \calP_{\Omega}(\x_n)$, where $\calP_{\Omega}$ is an operation that retains only the entries from the index set $\Omega$ and sets all the remaining entries to zero. Here, $\Omega$ contains the index of all known entries in the data. Since $\one^T\z_n = 1$, the last constraint for the missing entries is:
\begin{align}
\sum_{i\in\Omega^c}z_{in} = 1 - \sum_{m\in\Omega}x_{mn}, \forall n=1,\cdots,N
\end{align}
where $\Omega^c$ is the complement of $\Omega$.

With the above analysis, we now propose our model for expense forecast
\begin{equation}\label{eq:mc1}
    \begin{aligned}
    \min_{\W,\H,\Z} ~& \big\|\W\H^T - \Z\big\|_F^2 \\
    \mathrm{s.~t.}\quad 
    & \one^T\W = \one^T, \W\geq\boldsymbol{0}, \H\one = \one, \H\geq\boldsymbol{0} \\
    & \one^T\Z = \one^T, \Z\geq\boldsymbol{0}, \calP_{\Omega}(\Z) = \calP_{\Omega}(\X)
    %& z_{in} \geq 0,~\sum_{i\in\Omega^c}z_{in} = 1 - \sum_{m\in\Omega}x_{mn}, \\
    %&\forall n=1,\cdots,N
\end{aligned}
\end{equation}
%\reminder{A more compact way to state the last constraint is $\one^T\Z = \one^T, \Z\geq\boldsymbol{0}$, which also emphasizes the triple-simplex nature of the problem. For the algorithmic solution, the way you wrote the constraint is more convenient, but for conceptualizing the problem the above is better I think} 
where $\X=[\x_1,\cdots,\x_N]$ holds the (incomplete) input data, $z_{in} \geq 0,~\sum_{i\in\Omega^c}z_{in} = 1 - \sum_{m\in\Omega}x_{mn}$, and $\Z=[\z_1,\cdots,\z_N]$ holds the output of the algorithm. The model minimizes the squared Frobenius norm of the difference between $\W\H^T$ and $\Z$, subject to constraints on $\W$, $\H$, and $\Z$. The constraints on $\W$ and $\H$ enforce the low-rank approximation within the budget limits, while the constraints on $\Z$ ensure that the sum of the predicted and observed expenses is within the budget limit. %\footnote{"Within budget" limit means $\leq$ budget, but you actually enforce equality} 
%$\calP_{\Omega}$ is an operation that retains the entries from the index set $\Omega$ and sets the remaining entries to zero, and $\Omega$ is the set of known entries in the data.

\subsection{Optimization}
Although Problem \eqref{eq:mc1} is not a convex problem, if we fix any two variables in \eqref{eq:mc1}, the remaining variable's subproblem becomes convex. As a result, we use an inexact alternating optimization method to solve \eqref{eq:mc1}.

\subsubsection{Update $\W$}

Assuming we have estimated $\W$, $\H$, and $\Z$ after $r$ iterations, we need to solve the subproblem with respect to $\W$ for the next iteration, i.e.,
\begin{equation}\label{eq:minW}
    \begin{aligned}
        \min_\W ~& \big\|\W(\H^{(r)})^T - \Z^{(r)}\big\|_F^2 \\
        \mathrm{s.~t.}~ & \one^T\W = \one,~\W\geq\boldsymbol{0}
    \end{aligned}
\end{equation}
exact solution is expensive as it requires many iterations to converge. Hence, we use an alternative method, motivated by the Majorization-Minimization approach, which minimizes a tight upper bound of the loss function. To achieve this, we calculate a tight upper bound of $f(\W)$ by using a quadratic function $f(\W; \hat\W^{(r)}) = f(\hat\W^{(r)}) + \big\langle\nabla f(\hat\W^{(r)}), \W - \hat\W^{(r)}\big\rangle + \notag\\
    \frac{\gamma^{(r)}_w}{2}\|\W - \hat\W^{(r)}\|_F^2$,
% \begin{align}
%     f(\W; \hat\W^{(r)}) =&~ f(\hat\W^{(r)}) + \big\langle\nabla f(\hat\W^{(r)}), \W - \hat\W^{(r)}\big\rangle + \notag\\
%     &\frac{\gamma^{(r)}_w}{2}\|\W - \hat\W^{(r)}\|_F^2 \notag
% \end{align}
where $\gamma^{(r)}_w$ is a parameter associated with the step size which can be chosen as the spectral norm of $(\H^{(r)})^T\H^{(r)}$, $\langle \cdot \rangle$ is the inner product, the gradient $\nabla f(\hat\W^{(r)})$ is given by $\nabla f(\hat\W^{(r)}) = \left(\hat\W^{(r)}(\H^{(r)})^T - \Z^{(r)}\right) \H^{(r)}$, 
% \begin{equation}\label{eq:gradW}
%     \nabla f(\hat\W^{(r)}) = \left(\hat\W^{(r)}(\H^{(r)})^T - \Z^{(r)}\right) \H^{(r)}
% \end{equation}
and $\hat{\W}^{(r)} = \W^{(r)} + \beta^{(r)}(\W^{(r)} - \W^{(r-1)})$. We choose the extrapolation weight based on the following updating scheme 
\begin{equation}
    \begin{cases}
        \beta^{(r)} = (1 - t^{(r)}) / t^{(r+1)} \\
        t^{(r)} = \frac{1 + \sqrt{4(t^{(r-1)})^2+ 1}}{2},~\mathrm{with~} t^{0} = 0
    \end{cases}
\end{equation}
which comes from Nesterov’s accelerated gradient descent \cite{nesterov2013gradient}.

Now instead of solving \eqref{eq:minW} exactly, we propose to solve
\begin{equation}%\label{eq:minW2}
    \begin{aligned}
        \min_\W ~& f(\W; \hat\W^{(r)}) \\
        \mathrm{s.~t.}~ & \one^T\W = \one,~\W\geq\boldsymbol{0}
    \end{aligned}\notag
\end{equation}
which has a closed-form solution as
\begin{align}\label{eq:updatew}
    % \W^{(r+1)} = \mathcal{P}_{\mathcal{W}_n}\left(\W^{(r)} - \nabla f(\hat\W^{(r)}) / \gamma_w^{(r)}\right)
    \w_f^{(r+1)} = \mathrm{prox}_{\mathcal{W}_f}(\bar{\w}_f^{(r+1)}),~\forall f=1,\cdots,F
\end{align}
where $\mathrm{prox}_{\mathcal{W}_f}$ is the proximal operator that projects its argument onto a convex set $\mathcal{W}_f=\{\w_f\in\bR^M \mid \one\w_f = 1,~\w_f\geq\boldsymbol{0}\}$, and $\bar{\w}_f$ is the $f$-th column of 
$$\bar{\W}^{(r+1)} = \W^{(r)} - \nabla f(\hat\W^{(r)}) / \gamma_w^{(r)}.$$
Eq.~\eqref{eq:updatew} is closed-form and we refer to \cite{duchi2008efficient} for the solution of Euclidean projection of a vector onto the simplex.

\subsubsection{Update $\H$}
Since the subproblem w.r.t. $\H$ is the transpose of \eqref{eq:minW}, 
% \begin{equation}\label{eq:minH}
%     \begin{aligned}
%         \min_\H ~& \big\|\W^{(r+1)}(\H)^T - \Z^{(r)}\big\|_F^2 \\
%         \mathrm{s.~t.}~ & \H\one = \one,~\H\geq\boldsymbol{0}.
%     \end{aligned}
% \end{equation}
%The subproblem for optimizing $\H$ while keeping $\W$ and $\Z$ fixed is similar to the subproblem for optimizing $\W$ in \eqref{eq:minW}. The main difference is that the update for $\W$ is performed column-wise, while the update for $\H$ is performed row-wise since each row of $\H$ is constrained to lie on a probability simplex. \reminder{We can simply say that this subproblem is the transpose of the previous subproblem?} 
the detailed derivations are skipped, and the expression for updating the rows of $\H$ is presented as follows
\begin{align}\label{eq:updateh}
    \h_n^{(r+1)} = \mathrm{prox}_{\mathcal{H}_n}(\bar{\h}_n^{(r+1)}),~\forall n=1,\cdots,N
\end{align}
where $\bar\h_n^{(r+1)}$ is the $n$-th row of
$\bar{\H}^{(r+1)} = \H^{(r)} - \nabla f(\hat\H^{(r)}) / \gamma_h^{(r)}$ with $\nabla f(\hat{\H}^{(r)}) = (\hat\H^{(r)}(\W^{(r+1)})^T - (\Z^{(r)})^T)\W^{(r+1)}$ and $\hat\H^{(r)} = \H^{(r)} + \beta^{(r)}(\H^{(r)} - \H^{(r-1)})$.

\subsubsection{Update $\Z$}
The subproblem w.r.t. $\Z$ is 
\begin{equation}\label{eq:minZ}
    \begin{aligned}
        \min_\Z ~& \big\|\Z - \W^{(r+1)}(\H^{(r+1)})^T\big\|_F^2 \\
        \mathrm{s.~t.}~ & \calP_{\Omega}(\Z) = \calP_\Omega(\X), \one^T\Z = \one^T, \Z\geq\boldsymbol{0}.
        %&\sum_{i\in\Omega^c}z_{in} = 1 - \sum_{m\in\Omega}x_{mn}, \forall n=1,\cdots,N.
    \end{aligned}
\end{equation}
This problem can be split into two distinct subproblems related to the entries in $\Omega$ and $\Omega^c$, respectively. For the entries in $\Omega$, the updating rule is $z^{(r+1)}_{mn} = x_{mn},~\forall m,n \in \Omega$, 
% \begin{align}\label{eq:soluz1}
%     z^{(r+1)}_{mn} = x_{mn},~\forall m,n \in \Omega
% \end{align}
while for those in $\Omega^c$, the subproblem becomes an Euclidean projection onto a simplex defined as $\mathcal{Z}_n = \{z_{in},~\forall i,n\in\Omega^c \mid \sum_{i,n\in\Omega^c}z_{in} = 1 - \sum_{m,n\in\Omega}x_{mn}, z_{in}\geq 0\}$.
% The first subproblem is in terms of the entries in $\Omega$, i.e.,
% \begin{equation}\label{eq:minZ1}
%     \begin{aligned}
%         \min_{\{z_{mn}\}} ~& \sum_{m,n\in\Omega}\left(z_{mn} - \w_m^{(r+1)}(\h_n^{(r+1)})^T\right)^2 \\
%         \mathrm{s.~t.}~ & z_{mn} \geq 0, ~ z_{mn} = x_{mn}, ~\forall m,n\in\Omega
%     \end{aligned}
% \end{equation}
% where the solution is straitforward, i.e.,
% \begin{align}\label{eq:soluz1}
%     z_{mn}^{(r+1)} = x_{mn},~\forall m,n\in\Omega.
% \end{align}
% The second subproblem is related to the missing entries in $\Omega^c$. For the $n$-th project, its subproblem is written as
% \begin{equation}\label{eq:minZ2}
%     \begin{aligned}
%         \min_{\{z_{in}\}} ~& \sum_{i,n\in\Omega^c}\left(z_{in} - \w_i^{(r+1)}(\h_n^{(r+1)})^T\right)^2 \\
%         \mathrm{s.~t.} ~& z_{in} \geq 0,~\sum_{i\in\Omega^c}z_{in} = 1 - \sum_{m\in\Omega}x_{mn}
%     \end{aligned}
% \end{equation}
% which is a problem of Euclidean projection onto the simplex $\mathcal{Z}_n = \{z_{in},~\forall i,n\in\Omega^c \mid \sum_{i,n\in\Omega^c}z_{in} = 1 - \sum_{m,n\in\Omega}x_{mn}, z_{in}\geq 0\}$. 
Its solution can be written as
% % Defining
% % $\tilde{\z}_{n}^{(r+1)}=\mathrm{prox}_{\mathcal{Z}_n}\left( \W_{\Omega^c}^{(r+1)}\h_n^{(r+1)} \right)$, \reminder{Should it be proximal operator or simply projection operator here?} 
% % the solution of \eqref{eq:minZ2} is then expressed as
% \begin{align}\label{eq:soluz2}
%     z_{mn}^{(r+1)} = \tilde{z}_{mn}^{(r+1)},~\forall m,n\in\Omega^c
% \end{align}
$z_{mn}^{(r+1)} = \tilde{z}_{mn}^{(r+1)},~\forall m,n\in\Omega^c$, 
where $\tilde{z}_{mn}^{(r+1)}$ is the $m$-th element in $\tilde{\z}_{n}^{(r+1)}=\mathrm{prox}_{\mathcal{Z}_n}\left( \W_{\Omega^c}^{(r+1)}\h_n^{(r+1)} \right)$, and $\W_{\Omega^c}$ contains the rows of $\W$ that are corresponding to the missing expenses in project-$n$.
% Combining \eqref{eq:soluz1} and \eqref{eq:soluz2} yields 
With this analysis, $\z_n$ can be updated as
\begin{align}\label{eq:updatez}
    z_{mn}^{(r+1)} = 
    \begin{cases}
        x_{mn},~\forall m,n\in\Omega \\
        \tilde{z}_{mn}^{(r+1)},~\forall m,n\in\Omega^c.
    \end{cases}
\end{align}

It is important to note that the expense estimates obtained from \eqref{eq:updatez} represent the fractional expenses relative to the project's budget. In order to obtain the absolute expense estimates for each project, we must multiply $\z_n$ by its corresponding budget $B_n$:
\begin{align}\label{eq:soluy}
\hat{\y}_n = B_n\z_n^{(R)}.
\end{align}
Here, $\z_n=[z_{1n},\cdots,z_{Mn}]$ and $R$ represents the final iteration of the TSMC algorithm when it converges.
% \reminder{Why NOMIC? You never explain in the paper what the acronym means. Another idea could be TSMC: triple simplex matrix completion} 

We note that TSMC belongs to a family of block coordinate descent (BCD) algorithms. Since the original subproblem w.r.t. $\W$ or $\H$ is a convex constrained least squares problem, the prox-linear mapping is known to have a Lipschitz continuous gradient. The update of $\Z$ is also closed-form. Therefore, according to \cite{xu2013block}, TSMC converges to a stationary point.

The detailed steps of TSMC are summarized in Algorithm \ref{alg:proposed}. Its primary computational complexity is the calculation of the gradients of $\W$ and $\H$ and projecting them onto the probability simplex. The overall complexity of TSMC is $O(RMNF)$, where $R$ is the total number of iterations required for convergence.

\begin{algorithm}[t]
\small
\caption{The TSMC algorithm}\label{alg:proposed}
\begin{algorithmic}[1]
\State\textbf{Input} {$\{x_{ij},\forall i,j\}$, $F$}

\State Initialize $\W$ and $\H$ such that both matrices are non-negative, the rows of $\W$ sum up to one and the columns of $\H$ sum up to one.

\While{Stopping criteria has not been reached}
  % \For{$f = 1;\ f \leq F;\ f = f + 1$}
  %   Update the $f$-th column of $\W$ via \eqref{eq:updatew}
  % \EndFor
  
  % \For{$n = 1;\ n \leq N;\ n = n + 1$}
  %   Update the $f$-th row of $\H$ via \eqref{eq:updateh}
  % \EndFor

  % \For{$n = 1;\ n \leq N;\ n = n + 1$} 
  %   Update $\z_n$ via \eqref{eq:soluz1} and \eqref{eq:soluz2}
  % \EndFor

  Update $\W$ via \eqref{eq:updatew}

  Update $\H$ via \eqref{eq:updateh}

  Update $\Z$ via \eqref{eq:updatez}
  
\EndWhile

\State Estimate the expense estimates via \eqref{eq:soluy}

\State\textbf{Output} $\{\hat{\y}_1,\cdots,\hat{\y}_N\}$
%$, \W, \H$

\end{algorithmic}
\end{algorithm}
\section{Experiments}

In this section, we evaluate the performance of TSMC by comparing it with other cutting-edge algorithms, using two real expense datasets. The first dataset consists of the costs associated with investigators in 367 clinical trials spanning from December 2012 to March 2023. 
The second dataset covers pass-through expenses of 449 projects from August 2012 to March 2023. To simulate a practical scenario, both datasets are split into training and testing sets, with data before and after January 2021 being used for each set, respectively. Here, January 2021 is chosen to represent a prediction date for expenses beyond that point. To stack data points for each project into a matrix, the dates of every project are then converted into a month index, starting from 0 for the first date and increasing by 1 for subsequent dates. This creates 11,256 training samples and 5,727 testing samples for the first dataset, and 14,472 training samples and 7,354 testing samples for the second dataset.

The performance was evaluated using two metrics: root mean squared error (RMSE) and relative RMSE. RMSE measures the average error between the true expense values and their estimates, while relative RMSE takes into account the proportion of this error relative to the magnitude of the true values. These metrics are calculated based on the formulas: $\text{RMSE} = \|\x - \hat{\x}\|_F/\sqrt{N}$  and $\text{relative RMSE} = \|\x - \hat{\x}\|_F / \|\x\|_F$, where $N$ is the number of testing samples, $\x$ contains the true expense values, and $\hat{\x}$ is the corresponding estimate.
We compared TSMC with four baseline methods for matrix completion including the block successive upper bound minimization (BSUM) based NMC \cite{xu2013block}, convolutional Sparse Tensor Completion (CoSTCo) \cite{liu2019costco}, joint multi-linear and nonlinear identification (JULIA) \cite{qian2022julia}. In addition, we also include two missing value imputation algorithms, namely, K-nearest neighbor (KNN) \cite{troyanskaya2001missing} and median imputation, which are efficient for imputing missing values in a matrix through two Python modules KNNImputer and SimpleImputer, respectively. Based on the fact that the expenses are time-series, we included the AutoRegressive Integrated Moving Average (ARIMA) technique into our analysis. For this approach, every project within the testing dataset is predicted by its individual ARIMA model\footnote{It's important to note that we omitted any comparison involving Recurrent Neural Network (RNN) methodologies. This decision was primarily driven by the fact that a significant number of projects within the training dataset contain fewer than 10 data samples, an inadequate amount for effectively training deep learning models like RNNs.}.  In terms of hyper-parameters, we assigned a rank of 3 to TSMC, BSUM, and CoSTCo. For JULIA, we utilized 3 linear components and 2 nonlinear components. Additionally, we used $K=10$ nearest neighbors for KNN. 

To determine when to stop the methods, we used two criteria: either the decrease of the objective function between two iterations is less than $10^{-4}$; or the number of iterations reaches 100. We note that all baseline methods do not have the capability to limit their forecasts within a given project budget. As a result, we adjust their forecasts by scaling them appropriately to guarantee that the predicted costs for each undertaking correspond to the difference between the assigned budgets and the previously accrued expenditures.

%\reminder{What does this "error of the objective function between two iterations" mean? Do you mean, "the decrease of the cost function between two iterations is less than ..."?} 
%\qc{Fixed.}

% Table generated by Excel2LaTeX from sheet 'rank3'
\begin{table}[htbp]
  \centering
  \caption{Performance comparison}
    \begin{tabular}{lrr|rr}
    \toprule
          & \multicolumn{2}{c}{Investigator Fee} & \multicolumn{2}{c}{Pass-through} \\
    \midrule
    Method & RMSE  & Rel RMSE & RMSE  & Rel RMSE \\
    \midrule
    JULIA & 123208 & 0.7581 & 88579 & 0.7796 \\
    CoSTCo & 215668 & 1.3271 & 113624 & 1.0000 \\
    TSMC  & \textbf{103056} & \textbf{0.6341} & 84542 & 0.7441 \\
    BSUM  & 135756 & 0.8354 & 93150 & 0.8198 \\
    MEDIAN & 178829 & 1.1004 & 124542 & 1.0961 \\
    KNN   & 150384 & 0.9254 & 119836 & 1.0547 \\
    ARIMA & 122865 & 0.7560 & 87352 & 0.7688 \\
    \bottomrule
    \end{tabular}%
  \label{tab:rmse}%
  \vspace{-1em}
\end{table}%

\begin{figure}[t]
\minipage{0.48\linewidth}
  \includegraphics[width=1.1\linewidth]{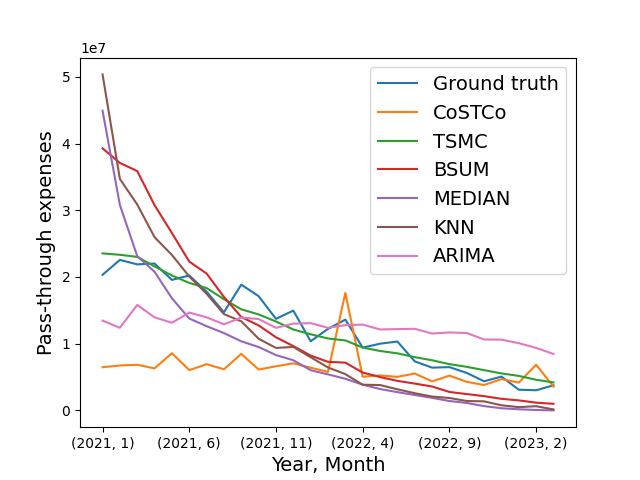}
  \caption{Predicted monthly investigator expenses.}\label{fig:inv}
\endminipage\hfill
\minipage{0.48\linewidth}
  \includegraphics[width=1.1\linewidth]{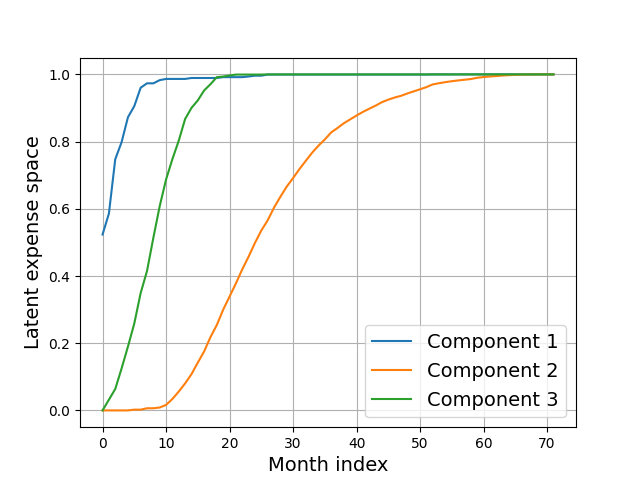}
  \caption{Expense patterns learned by TSMC, where curves show cumulative sums of the columns of  $\W$.}\label{fig:hest}
\endminipage
\end{figure}

The RMSE and relative RMSE results for estimating the expenses of pass-throughs and investigator fees are presented in Table~\ref{tab:rmse}. The results indicate that TSMC outperforms other methods on both datasets. This finding is further supported by Fig.~\ref{fig:inv}, which depicts the monthly investigator expenses predicted by aggregating all projects. It can be observed that the projections made by TSMC align closely with the actual expenses over time, while other baseline methods show varying levels of discrepancy.

In Fig.~\ref{fig:hest} shows three expense patterns TSMC learned from the investigator fee dataset. These patterns represent cumulative sums of the expense ratios in each column of $\W$ and reach 100\% when the budget is used up. Component 1 is the fastest pattern that uses up the budget in about a year, while Component 2 is the slowest, taking about 5 years. We can use the likelihood of a project belonging to different expense patterns, represented by each row of matrix $\H$, to cluster the projects. TSMC found that 14 projects are associated with Component 1, 216 with Component 2, and 137 with Component 3. Component 1 mostly consists of short Phase I studies with an average budget of \$424,718, while Component 3 mostly consists of longer Phase III studies with an average budget of \$2,022,607. This finding is consistent with the fact that Phase I studies typically last a few months, while Phase III studies can take a few years to complete. Additionally, projects with larger budgets generally require more time to complete than those with smaller budgets, which supports the interpretability of TSMC's findings.

\section{Conclusion}
This paper has introduced a novel technique called TSMC for addressing the challenge of project-level expense forecasting. TSMC is a triple simplex-constrained non-negative matrix completion method that can effectively deal with predicting expenses while adhering to budget constraints. In addition, it offers a high level of interpretability, allowing users to understand the factors that contribute to rapid or slow expenditure of the budget. The numerical results with real-life data  demonstrate that  TSMC outperforms the prior art in this task.

\bibliographystyle{IEEEtran}
\bibliography{reference}

\end{document}